\documentclass{article}

     \usepackage[final]{neurips_2024}

\usepackage[utf8]{inputenc} 
\usepackage[T1]{fontenc}    
\usepackage{hyperref}       
\usepackage{url}            
\usepackage{booktabs}       
\usepackage{amsfonts}       
\usepackage{nicefrac}       
\usepackage{microtype}      
\usepackage{graphicx}       
\usepackage{amsmath}
\usepackage{float}
\usepackage{algorithm}
\usepackage{algpseudocode}
\usepackage{color,soul}
\usepackage{colortbl} 
\usepackage[table,xcdraw]{xcolor} 
\usepackage[utf8]{inputenc}
\usepackage[moderate ]{savetrees}
\setul{0.2ex}{0.2ex}
\setulcolor{orange}
\newcount\Comments  
\newcommand{\kibitz}[2]{\ifnum\Comments=1\textcolor{#1}{#2}\fi}

\title{Proof Flow: Preliminary Study on Generative Flow Network Language Model Tuning for Formal Reasoning}

\author{%
  Matthew Ho \\
  UC San Diego\\
  La Jolla, CA\\
  \texttt{mah048@ucsd.edu}\\
  \And
  Vincent Zhu \\
  UC Santa Barbara\\
  Santa Barbara, CA\\
  \texttt{vincentzhu@ucsb.edu}\\
  \And 
  Xiaoyin Chen \\
  Mila, Université de Montréal\\
  Montréal, Québec, Canada\\
  \texttt{xiaoyin.chen@mila.quebec}\\
  \And
  Moksh Jain \\
  Mila, Université de Montréal\\
  Montréal, Québec, Canada\\
  \texttt{moksh.jain@mila.quebec}\\
  \And
  Nikolay Malkin \\
  University of Edinburgh\\
  Edinburgh, UK\\
  \texttt{nmalkin@ed.ac.uk}\\
  \And  
  Edwin Zhang \\
  OpenAI\\
  San Francisco, CA\\
  \texttt{edwin@openai.com}\\
}

\begin{document}

\maketitle

\begin{abstract}
\label{abstract}
  Reasoning is a fundamental substrate for solving novel and complex problems.
  Deliberate efforts in learning and developing frameworks around System 2 reasoning have made great strides, yet problems of sufficient complexity remain largely out of reach for open models. To address this gap, we examine the potential of Generative Flow Networks \citep[GFlowNets;][]{bengio_flow_2021, hu_amortizing_2024} as a fine-tuning method for LLMs to unlock advanced reasoning capabilities. In this paper, we present a proof of concept in the domain of formal reasoning, specifically in the Neural Theorem Proving (NTP) setting, where proofs specified in a formal language such as Lean can be deterministically and objectively verified. Unlike classical reward-maximization reinforcement learning, which frequently over-exploits high-reward actions and fails to effectively explore the state space, GFlowNets have emerged as a promising approach for sampling compositional objects, improving generalization, and enabling models to maintain diverse hypotheses. Our early results demonstrate GFlowNet fine-tuning's potential for enhancing model performance in a search setting, which is especially relevant given the paradigm shift towards inference time compute scaling and ``thinking slowly.'' 
  Code: \small \url{https://github.com/matt-seb-ho/gfn_ntp}
\end{abstract}

\section{Introduction}
Large language models (LLMs) have demonstrated impressive capabilities in pattern recognition and surface-level tasks, but still exhibit significant limitations in reasoning, particularly in complex logical inference and problem-solving tasks. Extending LLMs beyond mere memorization requires enhancing their reasoning abilities through approaches like System 2’s "slow thinking," which scales inference time computation to match problem complexity. 
A promising technique in this space is GFlowNet fine-tuning \citep{hu_amortizing_2024}, which unlocks new possibilities for search (a fundamental approach to reasoning \citep{SIMON19837}) by teaching the model to sample trajectories proportional to reward.
In this sense, fine-tuning model $M$ with the GFlowNet objective moves the inference time cost of sampling more suggestions (``slow thinking'') from $M$ to training time, thus amortizing the cost of inference.

Reasoning benchmarks like GSM8K \citep{cobbe_training_2021} and MATH \citep{hendrycks_measuring_2021} are increasingly subject to overfitting as models are trained specifically on these formats and similar math problems, raising concerns about their ability to capture the true generalization ability of models for real-world problem-solving. To address this, we turn to formal mathematics and neural theorem proving (NTP), leveraging proof assistants like Lean \citep{yang_leandojo_2023}. Built on dependent type theory, Lean can automatically apply common proof patterns through tactics—commands such as \texttt{intro, apply, simp}—which directly manipulate the proof state, transforming theorem proving into a formalized search problem. Unlike standard benchmarks, NTP offers interactive feedback and guarantees correctness, ensuring that flawed reasoning cannot yield correct results—a common issue in iterative self-teaching approaches \citep{zelikman_star_2022}.

In this paper, we present early in-progress evidence that GFlowNet fine-tuning has potential to accelerate search in challenging reasoning domains like theorem proving. \textbf{Our contributions include:}


(1) A study on GFlowNet fine-tuning’s promise for accelerating search in a domain challenging even for human experts. (2) An extensible code base integrating GFlowNet with the Lean environment.(3) An ablation study of key interventions in GFlowNet fine-tuning, including reward models and trajectory replay. (4) Early empirical results showing GFlowNet fine-tuning improves exploration and reasoning in neural theorem proving tasks.

\begin{figure}[!t]
    \centering
    \includegraphics[width=1\linewidth,trim={0, 0 0 0},clip]{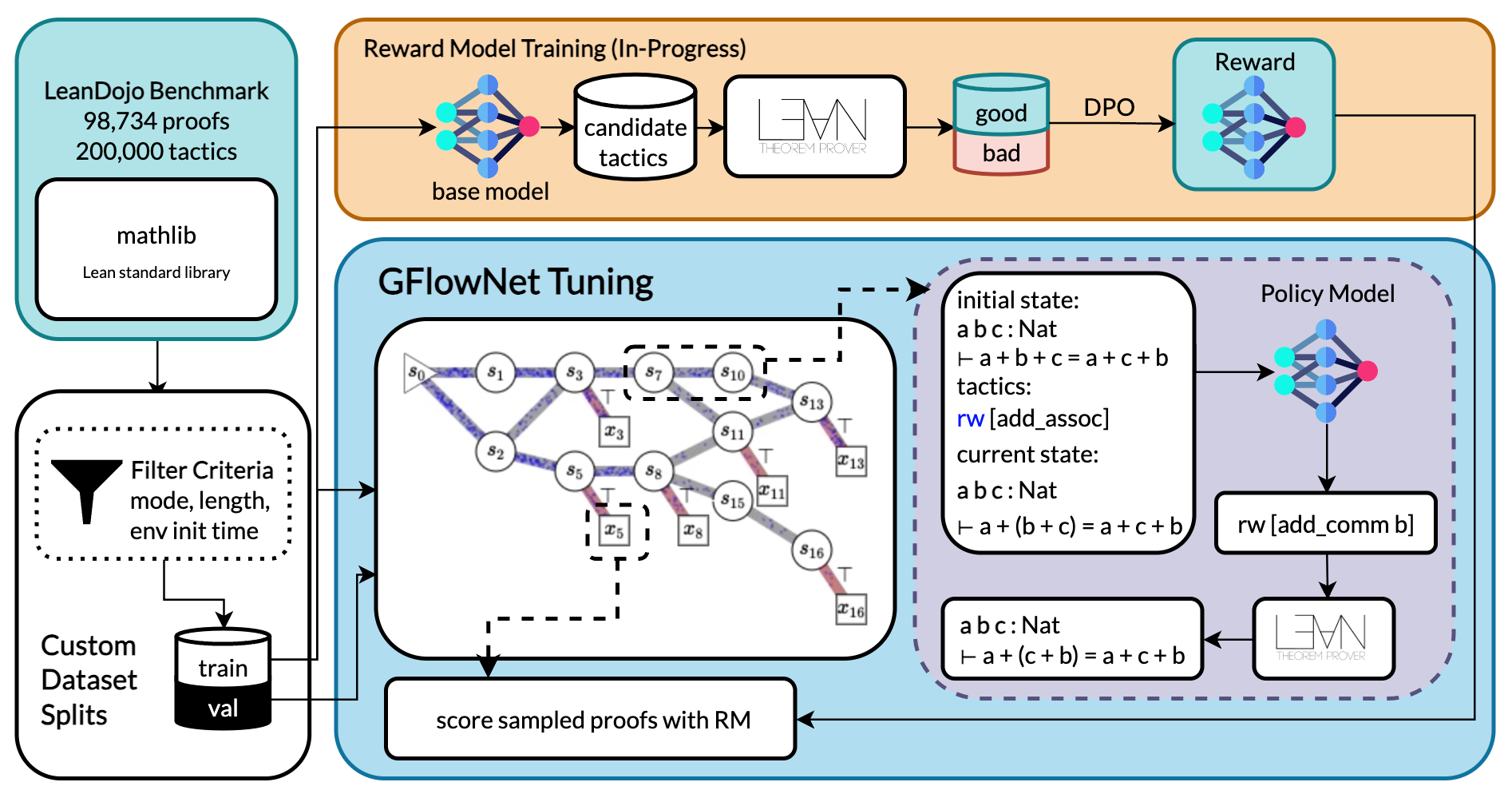}
    \caption{\textbf{Proof Flow System}. We extract and filter ground-truth proofs and theorems from LEAN 4's standard math library: mathlib. For reward model training data, we sample candidate tactics using a base model and label with Lean. We use Best First Search \citep{pearl1984heuristics} for evaluation. The GFlowNet diagram is a frame of an animation from \cite{bengio_GFlowNet_2022}}
    \label{fig:main-result}
    \vspace{-15pt}
\end{figure}

\section{Methodology}
\label{headings}

In the context of language modeling, GFlowNets are a maximum entropy RL algorithm for training policies to sample compositional objects with probability proportional to a reward~\citep{bengio_flow_2021,hu_amortizing_2024} (see preliminary details in \autoref{app:gflownet_overview}).
Following past NTP work (\autoref{app:related}), and leveraging the interactive features of Lean, we parameterize proof search as a tactic generation problem.
GFlowNets model the construction of the object as a terminating trajectory through a graph where edges (actions) specify adding a component to the object and nodes (states) are intermediate states of the object.
In this view, NTP can be viewed as composing proofs from component tactics. We refer the reader to \citet{hu_amortizing_2024} for a detailed discussion about GFlowNets in the context of language model fine-tuning.

We define our forward policy as $P_f(t |s)$ where $t$ is a tactic and $s$ is current proof state, all prior tactics taken, and initial proof state (in Lean, tactic states include the goal).
Since GFlowNet graphs allow nodes to have multiple parents (analogously, multiple Lean tactic sequences can lead to the same proof state), optimizing GFlowNet objectives require learning a backwards policy (specifying how reward arrived in current node from).
Not only would this be difficult as it requires learning the inverse of the function that the base model learned in fine-tuning, it would also add another source of training instability.
To avoid this problem, we use a state encoding that includes the trajectory's history.
Instead of representing a partially constructed proof by just its current proof state, we include its initial state and all previous tactics.
This enforces a tree structure, making the backward policy trivial by ensuring that each proof state has only one parent.

As for the reward, one option is to use the feedback provided by the Lean verifier as a binary reward.
While strictly correct and therefore un-hackable, this binary reward may be too sparse and penalize promising partial trajectories that were unable to complete due to limited search budget.
To address this, we introduce partial reward through a reward model (RM).
Here, we take advantage of ReProver's training objective 
(maximizing $P(t|\text{current\_proof\_state})$ for ground truth proof trajectories)
and use it to score the individual tactics in a partial trajectory (note that for the RM, we use the history-less state encoding, or just the current proof state). The trajectory balance~\citep[TB;][]{malkin_trajectory_2023} learning objective for GFlowNets also requires estimating the log partition function $\log Z$, where $Z$ is the sum of rewards for all terminal states for the given theorem.
To that end, for each theorem, we run a forward pass of the policy model on the initial state, and feed the model's final hidden states to a linear layer that predicts $\log Z$.
This linear layer is learned simultaneously with the policy.

The trajectories used for training are sampled from the current policy in conjunction with the Lean environment to verify syntactic correctness and yield next states. In additon, we sample ground truth correct trajectories which we are given access to during train time, whereas during the test time evaluation and validation loops, we use standard best first proof search. Please see \autoref{app:further-methodology} for more methodological details and detailed algorithm pseudocode.

\section{Experiments}

\label{sec:setup}

\textbf{Data.} Our experiments are based on the LeanDojo benchmark \cite{yang_leandojo_2023}. 
Starting from the Lean4 \texttt{random} splits version, we apply several rounds of filtering to form our train and validation splits (see \autoref{app:data_filter} for details). The final train and validation splits contain 1K and 20 instances, respectively.~
\begin{figure}
\begin{minipage}[t]{.49\textwidth}
    \centering
    \includegraphics[width=\linewidth]{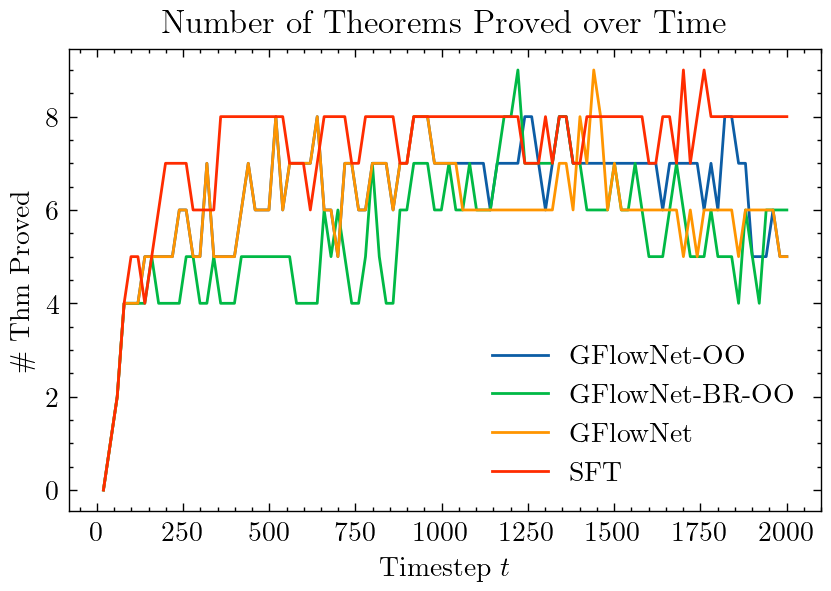}
    \caption{Evaluation on hold-out set of 20 theorems unseen during train time. Validation run every 20 gradient steps.  GFlowNet-OO refers to Online Only, GFlownet-BR-OO refers to Binary Reward and Online Only, while GFlowNet refers to the full method. SFT refers to Supervised Fine-Tuning, or just maximizing log likelihood on the ground truth trajectory. For exact theorem names and lengths see \autoref{tab:accuracy_results}.}
    \label{fig:num_proved}
\end{minipage}\hfill
\begin{minipage}[t]{.49\textwidth}
    \centering
    \includegraphics[width=\linewidth]{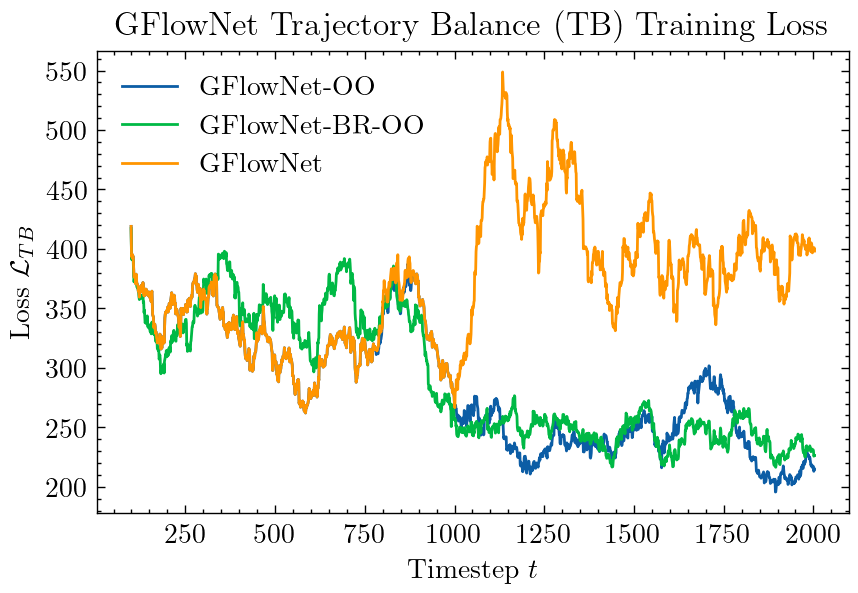}
    \caption{Training Trajectory Balance loss over 2000 gradient steps. Loss is smoothed using simple moving average over 100 steps. Interestingly, GFlowNet observes instability in training around episode 1000. Note that the full GFlowNet is the only off-policy method, as opposed to the on-policy GFlowNet-OO and GFlowNet-BR-OO.}
    \label{fig:training-loss}
\end{minipage}
\end{figure}
\textbf{Model.}
We conduct experiments initializing our model with ReProver, \cite{yang_leandojo_2023} a pretrained tactic generator, which was in turn initialized from ByT5-small \citep{xue2022byt5tokenfreefuturepretrained} a 350M parameter sequence-to-sequence model whose byte level vocabulary makes it well suited for the unicode-symbol-heavy Lean language. We train using a single A100 40GB GPU. Please see all other hyperparameters in \autoref{app:hparams}. \textbf{Ablations.} We conduct ablations over the reward function and replay buffer.
In standard setting (GFlowNet)---closest to the original GFlowNet LLM fine-tuning \citep{hu_amortizing_2024} setup---each training step samples from the replay buffer with probability 0.5.
In the online-only (GFlowNet-OO) setting, the replay buffer is unused-- every training step receives reward from a newly sampled trajectory.
In the binary-reward and online-only (GFlowNet-BR-OO) ablation, we use a binary reward that gives perfect score to correct trajectories and a length-penalized score to all other outcomes. 
In all GFlowNet runs, we inject the ground truth trajectory as way to help stabilize training, This particular setting makes GFlowNet most comparable to SFT, which we also include as a baseline.
Ongoing experiments explore removing this intervention and scaling train time exploration. \textbf{Compute Budget.} Proof search evaluation used much tighter constraints both due to time/resource limits, but also to test the model's efficiency in search.
Compared to ReProver's evaluation setting with search budget of 10 minutes and branching factor of 64, we use 30 seconds and branching factor 8. For the fairest comparison, the base model was evaluated using ReProver's original state encoding instead of the history-augmented encoding used for SFT and GFlowNet training runs.

\textbf{Preliminary Results Analysis.}
The main results are displayed in \autoref{tab:accuracy_results}.
We observe that under low resource constraints, GFlowNet fine-tuning is capable of enhancing proof search performance, as evidenced with the substantially improved solve rate compared to the base model. However, while promising, we note that the baseline method of Supervised Fine-Tuning is able to match or exceed the GFlowNet fine-tuning in solve rate, pointing towards the necessity for larger compute and further research. The GFlowNet ablations yield largely similar results to the full method. However, both ablations GFlowNet-OO and BR-OO slightly under-perform the full method up to timestep $1000$.
Interestingly, there is nearly no difference in the loss or performance between OO and BR-OO before step $1000$, which is likely due the fact that the RM is only used when a tactic is syntactically correct, which the model fails to generate before step $1000$.

While GFlowNet and GFlowNet-BR-OO settings achieve a higher peaks at $9$ theorems solved, they occur roughly two-thirds of the way into their runs before regressing through the end of training.  One possible explanation is that through the binary reward punishing incomplete trajectories, it discourages exploration and therefore harms test set generalization in more training steps. In addition, the full off-policy GFlowNet experiences training instability, which motivaties future research into instability mitigation. Some potential avenues may include distribution correction \citep{kumar2020discor} or improved replay buffer sampling strategies \citep{oh2021learningsamplelocalglobal}.
Indeed, prior works in GFlowNet fine-tuning have shown that replay buffer usage is essential to good runs in practice and our experiment at least confirms that training with replay was substantially faster, completing 2000 steps in .86
the time other runs took.
Thus while replaying off-policy proofs introduces instability in the short run, the training speedup is likely worth the tradeoff given strategies to stabilize the off-policyness.

For our low compute budget training and evaluation, SFT yielded better results than GFlowNet.
SFT has time and again been demonstrated to be an extremely efficient learning algorithm, and our setting is no exception. We hypothesize that GFlowNet fine-tuning's potential for improving exploration is not fully realized under our current constraints due to insufficiently well-performing prior and reward models, motivating further work in improving both components. Sampling diverse positive samples from the prior is critical for leveraging the GFlowNet objective to sample more diverse samples than a mode-seeking objective such as Policy Gradient \citep{NIPS1999_464d828b}, which our existing model is unlikely to do under our compute constraints. Thus, scaling the batch size and search budget during both train and inference time will likely lead to more pronounced gains and improvements over the SFT baseline.


\vspace{-5pt}
\section{In-Progress and Future Work, Limitations, and Conclusion}
\textbf{In-Progress and Future Work.} Further variations of the existing setting are currently being tested. 
In addition to continuing to search certain key hyperparameters (e.g. batch size, number of training steps, replay frequency, reward scaling, etc.), there are more involved interventions that are also currently in progress related to the reward model.
As GFlowNet fine-tuning explicitly aims to sample in proportion to reward, our method is gated by the quality of our reward model (reward model engineering details in \autoref{app:future-work}).
Additionally, other search approaches for exploration during training may be another fruitful direction, such as Monte-Carlo Tree Search (MCTS). As for future work, one important direction is extending GFlowNet fine-tuning across other formal reasoning tasks, moving towards a ``universal reasoner'' that generalizes across domains. This could involve tasks beyond Neural Theorem Proving (NTP), like probabilistic programming or program synthesis. Furthermore, GFlowNets' ability to generate multiple samples could help quantify uncertainty in reasoning. The GFlowNet framework also holds promise for more complex processes, such as structured chains of thought and long-term planning, offering new possibilities for LLMs in formal and informal environments. In addition, the scaling laws of System 2 reasoning performance with respect to amount of compute budget needs to be analyzed, and how to allocate that budget most efficiently amongst inference search, training search, and base model size.

\textbf{Limitations and Conclusion.} Our experiments, constrained by search budget during training and dataset size, may limit GFlowNet fine-tuning’s full potential. Larger datasets and longer training could further amplify its benefits, especially for larger models (stronger priors could find multiple high reward regions in training more easily). Efficient proof state exploration also remains a challenge. We hypothesize GFlowNet objectives for fine-tuning LLMs can serve as a more principled approach for improving exploration in the context reasoning tasks. 
Our early results on NTP show promise for GFlowNet fine-tuning even under tight train and search budgets.
While SFT performs similarly or even better in these low resource regimes, the underlying principles anticipate better results for GFlowNet fine-tuning with scale and we hope to motivate future work towards achieving such a result.

\clearpage
\bibliographystyle{abbrvnat}
\bibliography{main.bib}

\clearpage
\appendix

\section{Further Experimental Details and Hyperparameters}\label{app:hparams}

We conduct full fine-tuning over $2$ epochs, or $2$K gradient steps with AdamW optimizer, at lr \texttt{1e-4} with $0.5$ gradient clipping norm, and batches containing $5$ sampled trajectories along with the ground truth trajectory.
For online trajectories, we temper the generation with probability $0.666$ where temperature is uniformly sampled between $0.25$ and $1.0$.
Log reward formulated as shown below, where $\tau$ is a trajectory containing states $s_i$ and tactics $t_i$, $\alpha=8$, $c$ is max tactic length (88), $l = \frac{1}{n} \sum_{i=1}^n \text{len}(t_i)$, and $p_{RM}$ is the reward model.
\[
    \texttt{LogR}(\tau) = \begin{cases}
        0 & \text{if $\tau$ completes the proof},\\
        -15 + \alpha \ln{\frac{c - l}{c}} & \text{if $\tau$ ends with LeanError},\\
        \sum_{i=1}^n \frac{1}{\text{len}(t_i)} p_{RM}(t_i|s_i) & \text{otherwise}
    \end{cases}
\]

In the binary reward ablation, reward is instead formulated as:
\[
    \texttt{BinaryLogR}(\tau) = \begin{cases}
        0 & \text{if $\tau$ completes the proof},\\
        -15 + \alpha \ln{\frac{c - l}{c}} & \text{otherwise}\\
    \end{cases}
\]

\section{Related Works}
\label{app:related}
\subsection{Neural Theorem Proving}
Several approaches have been explored in the past to enhance the reasoning abilities of LLMs within formal mathematics.
Given the high degree of specialization in formal theorem proving and the relative recency of proof assistants, NTP can be considered a low-data domain especially compared to the related but less verifiable wealth of generic coding data.
As such, Reinforcement Learning (RL) has been natural approach to work around this constraint as well as leverage the interactivity of proof assistants.
Several works have emulated an AlphaZero-like approach of combining MCTS with online training methods such as policy gradients including TacticZero \citep{wu2021tacticzerolearningprovetheorems} and Hypertree Proof Search \citep{lample2022hypertreeproofsearchneural}.
Most recently, DeepSeek-Prover v1.5 \cite{xin_deepseek-prover-v15_2024} trained with GRPO \citep{shao2024deepseekmathpushinglimitsmathematical}, an alternative to PPO \cite{schulman_proximal_2017} that uses trajectory group rewards to circumvent the need to train a critic model.
While promising, these previous works share the long-established objective of RL algorithms: reward maximization. 
This regime has been shown to encounter issues such as reward hacking and mode collapse \citep{casper_open_2023}, hindering the learned model’s ability to generalize and find novel trajectories as it concentrates probability around the max reward trajectories found in training.

Another major technique employed to combat the low data environment is data augmentation.
Synthetic data is deployed in two orthogonal directions.
First, works like DeepSeek-Prover v1 \citep{xin2024deepseekproveradvancingtheoremproving} introduce new formal math data points by autoformalizing informal (natural language) high school and undergraduate math solutions scraped from the internet.
Secondly, works gather further training tokens by augmenting existing theorems and proofs with natural language ``thought'' annotations.
TheoremLlama \citep{wang_theoremllama_2024}, Lean-STaR \cite{lin_lean-star_2024}, and DeepSeek-Prover v1.5 \citep{xin_deepseek-prover-v15_2024} each leverage contemporary base models' improved understanding of natural language and their ability to perform Chain-of-Thought reasoning by injecting natural language explanations between proof steps.
These methods have achieved significant improvements, but in each case, the synthetic data is generated from some model, causing the method to be ultimately bottlenecked by the capability of said model.

\section{Overview of Generative Flow Networks}
\label{app:gflownet_overview}
Generative Flow Networks are a class of probabilistic models designed to sample complex, structured objects through sequential decision-making processes. Unlike traditional generative models or reinforcement learning approaches that aim to maximize expected rewards, GFlowNets aim to sample objects such that the probability of generating a particular object $x$ is proportional to a predefined non-negative reward function $R(x)$. This property makes GFlowNets particularly suited for tasks that require diverse exploration of high-reward regions in the state space.

\textbf{Flow Conservation and State Flows.} The core principle of GFlowNets is the conservation of probability flow through the state space. Each state $s$ has an associated positive flow value $F(s)$, representing the total "probability mass" passing through that state.

\pagebreak

\textbf{Flow Conservation Equation.}

For non-terminal states $s \notin \mathcal{S}_F$:
\[
F(s) = \sum_{s' \in \text{Ch}(s)} F(s') P_B(s\,|\,s')
\]
where:
\begin{itemize}
    \item $\text{Ch}(s)$ is the set of child states reachable from $s$.
    \item $P_B(s\,|\,s')$ is the backward policy, defining the probability of transitioning from $s'$ to $s$.
\end{itemize}

For terminal states $s \in \mathcal{S}_F$:
\[
F(s) = R(s)
\]
This ensures that the incoming flow equals the outgoing flow at each state, preserving the total probability mass across the network \citep{bengio_GFlowNet_2022}.

\textbf{Detailed Balance Condition.}
An alternative formulation is the detailed balance condition:
\[
F(s) P_F(s'\,|\,s) = F(s') P_B(s\,|\,s')
\]
where $P_F(s'\,|\,s)$ is the forward policy, defining the probability of transitioning from $s$ to $s'$.

This condition ensures consistency in the flow of probability between states in both forward and backward directions.

\textbf{Flow Conservation and State Flows.}

\textbf{Forward Policy ($P_F(s'\,|\,s)$).} The forward policy guides the generation of new states (tactics) from the current state:
\[
P_F(s'\,|\,s) = \frac{F(s') P_B(s\,|\,s')}{F(s)}
\]
This equation derives from the detailed balance condition and ensures that the forward transitions are consistent with the flow values.

\textbf{Backward Policy ($P_B(s\,|\,s')$).} The backward policy is typically defined based on the problem structure and can often be chosen to simplify computations. In the NTP setting, $P_B(s\,|\,s')$ could represent the likelihood of reverting a tactic or the inverse of a forward action.

\textbf{Trajectory Probabilities and Sampling.} The probability of a trajectory $\tau$ under the forward policy is: $P_F(\tau) = \prod_{t=0}^{n-1} P_F(s_{t+1},|,s_t)$. The induced distribution over terminal states $s_n$ satisfies: $P_F(s_n) = \frac{R(s_n)}{Z}$, where $Z$ is the partition function: $Z = \sum_{s \in \mathcal{S}_F} R(s)$. This means that the probability of sampling a particular proof is proportional to its reward, aligning the sampling process with the objective of exploring high-reward proofs

\textbf{Learning Objectives.} The primary goal is to learn the forward policy $P_F(s'\,|\,s; \theta)$, parameterized by $\theta$, such that the induced distribution over terminal states matches the target distribution defined by the reward function.

\textbf{Trajectory Balance Loss.} We employ the Trajectory Balance (TB) loss \citep{malkin_trajectory_2023} to train the model:
\[
\mathcal{L}_{\text{TB}}(\theta) = \mathbb{E}_{\tau \sim P_F} \left[ \left( \log R(s_n) - \log Z + \sum_{t=0}^{n-1} \log P_F(s_{t+1}\,|\,s_t; \theta) - \sum_{t=1}^{n} \log P_B(s_{t-1}\,|\,s_t) \right)^2 \right]
\]
This loss function encourages the model to produce trajectories whose cumulative log-probabilities match the log-reward of the terminal state, adjusted by the partition function. Moreover, TB loss provides better credit assignment along trajectories, leading to more stable and efficient training. This stability is contrasted with traditional reinforcement learning methods, which often suffer from high variance in gradient estimates due to sparse or delayed rewards. The TB loss mitigates this issue by incorporating the reward function directly into the loss, enabling smoother gradients and more consistent updates during training.

\pagebreak

\section{Further Methodology Details}\label{app:further-methodology}

We give our algorithm pseudocode in detail here. \texttt{SampleTactic(policy, state)} simply samples a model completion for prompt=state using some temperature.

\begin{algorithm}
\caption{GFlowNet Fine-Tuning Algorithm with TB Loss within NTP Setting}\label{gfn_algorithm_rm}
\begin{algorithmic}[1]
\Require Forward Policy model (policy), Replay Buffer Usage probability $p$, Theorem Dataset $\mathcal{D}$, \texttt{ReplayForward}($\tau$): gets log probability of old trajectory with current policy, \texttt{SampleTrajectory}(theorem, \texttt{RM}): samples a proof trajectory and computes $(\log p_f, \log r)$, num trajectories
\State Initialize \texttt{LeanDojo} environment for theor˛em proving
\Repeat
    \For{each theorem $\mathrm{Thm} \in \mathcal{D}$}
        \State batch $\gets$ []
        \If{random number $\in [0, 1] < p$}
            \For{trajectory $\in$ [1, 2, ..., \text{num trajectories}]}
                \State Select a trajectory $\tau = \{s_0 \xrightarrow{} ... \xrightarrow{} s_n\}$, $\log r$ from replay buffer $\mathcal{B}$
                \State $\log p_f \gets$ \texttt{ReplayForward}($\tau$)
                \State batch $\gets$ batch + ($\log p_f$, $\log r$, $\tau$)
            \EndFor
        \Else
            \For{trajectory $\in$ [1, 2, ..., \text{num trajectories}]}
                \State $(\log p_f, \log r, \tau) \gets$ \texttt{SampleTrajectory}($\mathrm{Thm}$\texttt{.initial\_state}, policy)
                \State batch $\gets$ batch + ($\log p_f$, $\log r$, $\tau$)
            \EndFor
            
        \EndIf
        \State Add batch to replay buffer $\mathcal{B}$
        \State Compute Trajectory Balance Loss: $\mathcal{L}_{\text{TB}}(\text{batch})$
        \State Update model parameters: $\theta \gets \theta - \eta \nabla_\theta \mathcal{L}_{\text{TB}}$
    \EndFor
\Until $\mathcal{L}_{\text{TB}}$ converged
\end{algorithmic}
\end{algorithm}
\begin{algorithm}
\caption{Sample Trajectory Procedure}
\label{gfn_sample_trajectory_algo}
\begin{algorithmic}[1]
\Require $\text{initial state, } \text{policy, } \text{max depth, }  \texttt{ComputeLogR, } \texttt{SampleTactic}$
\State tactics $\gets$ []
\State states $\gets$ [initial state]
\State state $\gets$ (initial state, tactics, initial state)
\State logprob $\gets$ 0
\For{step $\in [1, 2, ..., \texttt{max\_depth}]$}
    \State tactic, $\log p_f$ $\gets$ \texttt{SampleTactic}(policy, state)
    \State logprob $\gets$ logprob $+ \log p_f$
    \State next state $\gets$ Lean run tactic on state
    \State states $\gets$ states + [next state]
    \State tactics $\gets$ tactics + [tactic]
    \State state $\gets$ (initial state, tactics, next state)
\EndFor
\State return logprob, \texttt{ComputeLogR}(states, tactics), tactics
\end{algorithmic}
\end{algorithm}


\subsection{Dataset Filtering}
\label{app:data_filter}
The filtering criteria includes the following:
(1) Proof Style: Lean's tactic mode can be interwoven with standard term mode. For our experiments we only consider proofs that are entirely driven by tactics.
(2) Proof Length: for practical and generalization purposes, we examine limited trajectories and only consider theorems where the ground truth proof requires three or fewer steps
(3) State and Tactic length: memory constraints limited our study to theorems where the ground truth proof has state and tactic lengths under a threshold (under the ReProver tokenizer: 900 state tokens, 90 tactic tokens)
(4) Dojo Initialization Time: LeanDojo wraps a Lean REPL with a thorough dependency management system. We consider only theorems whose environment can be initialized within 5 seconds.
Due to time and resource constraints, our early experiments are run on a small subset of \texttt{mathlib} theorems. 
Our train set includes 1000 theorems with an extremely small validation split containing 20 theorems. 
Both contain a uniform distribution of proof lengths.

\subsection{Addressing Mode Collapse with GFlowNet-Based Sampling}
One of the central challenges of reinforcement learning in proof search is the occurrence of mode collapse \citep{casper_open_2023}, where the model prematurely converges on a small set of tactics and fails to adequately explore alternative solutions. This significantly hampers the model’s ability to generalize, as it becomes locked into a narrow trajectory of proof strategies. By using GFlowNets, we can ensure that the model continues to explore diverse proof strategies by assigning probabilities to entire proof sequences based on their compositional reward.

In contrast to temperature sampling, where randomness is injected at the token level, GFlowNets allow for sampling from the actual sequence distribution \citep{yu_flow_2024}, grounded in the likelihood of success in the proof domain. This approach both mitigates the risk of mode collapse and establishes a more structured investigation schema. The probability assigned to each tactic is directly informed by the reward associated with that tactic’s effectiveness in driving the proof forward, making the sampling process more principled and aligned with the task’s objectives.

\subsection{Amortizing LLM Inference for Efficient Sampling}
System 2 reasoning systems generally rely on increased inference time compute.
This typically manifests in using more tokens or searching more candidates.
The improved exploration efficiency of GFlowNets can also be understood as ``amortized LLM inference''  \citep{hu_amortizing_2024}.
Specifically, there are certain intractable posteriors such as $P(Y|X)^{1/T}$ (typically approximated using token-wise tempering), or in our case sampling proofs proportional to their reward.
We can use Monte Carlo methods to approximate sampling this, but through training with the GFlowNet objective, we effectively replace the compute spent on inference time Monte Carlo sampling to GFlowNet fine-tuning's train time, which consequently gets amortized over every inference.

\begin{table*}
\centering

\small  
\begin{tabular}{|c|l|c|c|c|c|c|}
\hline
Length & Theorems                            & Base & SFT & GFN-BR-OO & GFN-OO & GFN \\ \hline
             & Finset.card\_insert\_of\_not\_mem     &      &     &     &     &      \\ \cline{2-7}& Part.map\_map                       &   \cellcolor{green}   &     \cellcolor{green}& \cellcolor{green}    &      \cellcolor{green}& \cellcolor{green}      \\ \cline{2-7}
1& Ideal.isCompactElement\_top          &      &     &     &     &      \\ \cline{2-7}& CategoryTheory.Iso.trans\_conjAut      &   \cellcolor{green}   &     \cellcolor{green}&     \cellcolor{green}&      \cellcolor{green}& \cellcolor{green}      \\ \cline{2-7}
             & List.Nodup.erase\_get          &      &     &     &    &       \\ \cline{2-7}
             & Nat.dist\_eq\_sub\_of\_le        &      &     \cellcolor{green}&     \cellcolor{green}&     \cellcolor{green}&      \cellcolor{green}\\ \cline{1-7}& div\_nonneg                         &      &     \cellcolor{green}&     \cellcolor{green}&   \cellcolor{green}&   \cellcolor{green}     \\ \cline{2-7}
             & SimpleGraph.commonNeighbors\_top\_eq   &   \cellcolor{green}   &     \cellcolor{green}&     \cellcolor{green}&      \cellcolor{green}& \cellcolor{green}      \\ \cline{2-7}
             & zero\_le\_four                      &      &     &     &    &       \\ \cline{2-7}
2            & CategoryTheory.exact\_kernel  &      &     &     &    &       \\ \cline{2-7}
             & Matroid.Restriction.finite       &      &     &     &    &       \\ \cline{2-7}
             & Even.sub\_odd                       &      &     \cellcolor{green}&     \cellcolor{green}&   \cellcolor{green}&   \cellcolor{green}     \\ \cline{2-7}
             & Batteries.RBNode.Ordered.setRed     &   \cellcolor{green}   &     \cellcolor{green}&     \cellcolor{green}&     \cellcolor{green}& \cellcolor{green}      \\ \cline{1-7}& List.lookupAll\_length\_le\_one      &      &     &     &    &       \\ \cline{2-7}
             & Real.hasDerivAt\_negMulLog      &      &     &     &    &       \\ \cline{2-7}
             & Real.Angle.expMapCircle\_add        &      &     &     &    &       \\ \cline{2-7}
3            &Int.ediv\_two\_mul\_two\_add\_one\_of\_odd      &      &     \cellcolor{green}&     \cellcolor{green}&    &  \cellcolor{green}     \\ \cline{2-7}
             & Real.log\_of\_pos                  &      &     \cellcolor{green}&     \cellcolor{green}&    \cellcolor{green}&      \cellcolor{green}\\ \cline{2-7}
             & IsCompactlyGenerated.Boolean... &      &     &     &    &       \\ \cline{2-7}
             & Complex.natCast\_cpow\_natCast\_mul  &      &     &     &    &       \\ \hline
             Total& & 4/20& 9/20& 9/20 & 8/20 & 9/20\\ \hline
\end{tabular}

\caption{\textbf{Max Validation Results.} We ran proof search evaluation every 20 training steps. This table records the theorems proved in run's best solve rate validation step. Green cells indicates a proof was found under the search budget described in \autoref{sec:setup}}
\label{tab:accuracy_results}
\end{table*}

\subsection{Scalability of Inference Time Computation}
The power of scaling inference time compute has been recently been demonstrated with o1 \citep{openai_gpto1_2024}.
An additional benefit of the GFlowNet approach is its compatibility with this new paradigm.
By amortizing inference, we effectively reduce the computational cost of sampling additional proof strategies, making it feasible to explore a larger portion of the proof space within practical time limits \citep{hu_amortizing_2024}.

\subsection{Supervised Fine-Tuning (SFT) Training Loss}
\label{sft_train_loss}

\begin{figure}
\begin{minipage}[h]{.49\textwidth}
    \centering
    \includegraphics[width=\linewidth]{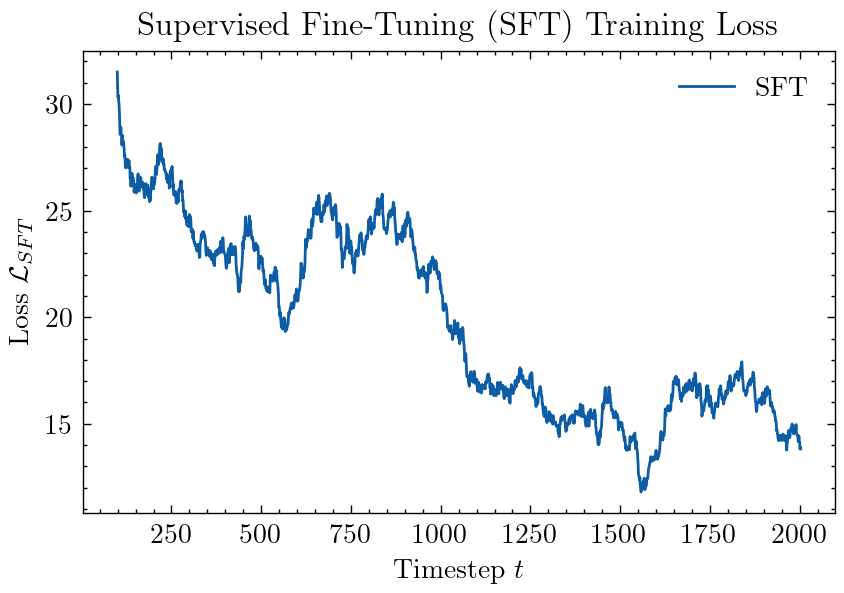}
    \caption{SFT Baseline Train Loss over Time}
    \label{fig:sft_train_loss}
\end{minipage}\hfill
\begin{minipage}[h]{.49\textwidth}
    \centering
    \includegraphics[width=\linewidth]{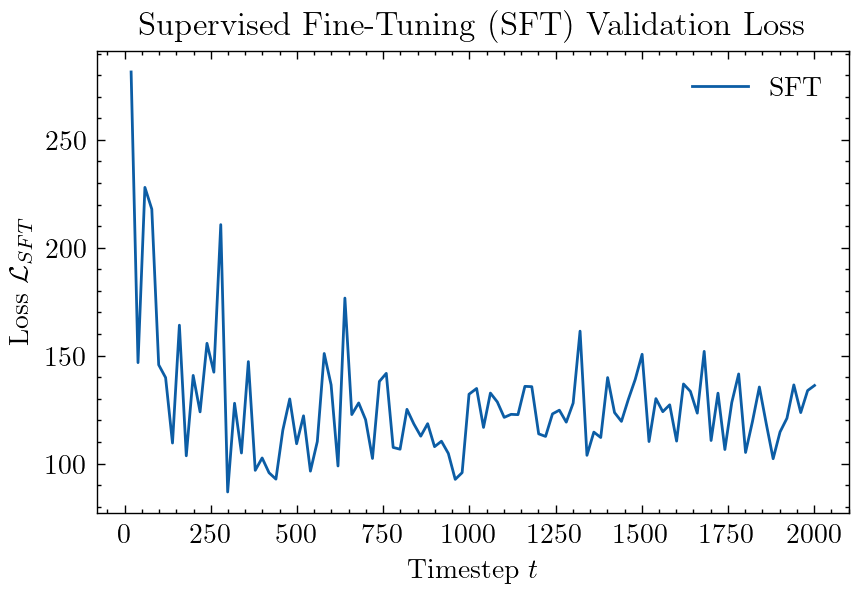}
    \caption{SFT Baseline Validation Loss over Time}
    \label{fig:sft_val_loss}
\end{minipage}
\end{figure}

The training loss curve for the Supervised Fine-Tuning (SFT) baseline illustrates steady convergence over the course of training, as seen in \autoref{fig:sft_train_loss}. The model starts with relatively high loss values, reflecting the initial difficulty in predicting the correct proof tactics. However, as training progresses, the loss steadily decreases, indicating that the model is learning to generate more accurate tactic sequences.

The loss function used for SFT is the standard cross-entropy loss, which measures the difference between the model’s predicted probability distribution for each tactic and the actual distribution from the ground truth. A major advantage of SFT is its stability and efficiency, as it does not suffer from the variance typically associated with reinforcement learning-based approaches, such as GFlowNet fine-tuning or PPO.

The validation loss, shown in \autoref{fig:sft_val_loss}, follows a similar trend to the training loss, with gradual improvement over time. This indicates that the model is not overfitting to the training data, but rather learning generalizable patterns. The gap between training and validation loss remains small throughout the process, which supports the observation that SFT, while not necessarily improving exploration, remains a highly effective method for stabilizing learning.

\section{In-Progress Work}\label{app:future-work}

\subsection{Reward Model Engineering} 

In our empirical analysis, we employed a base reward model (RM) as the verifier. However, in line with recent advancements in informal mathematical verifier training \citep{hosseini_v-star_2024}, we are working towards integrating more robust reward models. We have established a comprehensive pipeline for reward model training, consisting of data collection, data filtering, proof sampling, verification, training algorithms, and reward model evaluation. For training, we initially adopted a Supervised Fine-Tuning (SFT) approach, a well-established method for training language models. Additionally, we explored Direct Preference Optimization (DPO) \citep{rafailov_dpo_2023}, a technique that has gained significant attention in recent years.

While we could trivially use randomly shuffled tactics as negative examples for DPO, this would represent ``easy'' negatives where simple pattern matching could likely yield a decent performance.
To avoid learning shallow but ultimately unhelpful patterns, we prioritize gathering ``hard'' negatives.
By using the base reward model to sample proof trajectories and collecting tactics from failed proofs (as verified by Lean) we can identify incorrect tactics that the model assigns high likelihood-- constituting a somewhat on-policy approach.
One challenge in this data collection process is in annotating tactics.
Lean directly provides a trajectory-level annotation, but there is ambiguity in what the incorrect tactic could be.
To solve this, we need some degree of exploration.
To determine if tactic $t$ which yields state $s$ is truly a negative tactic, we can explore $s$ and determine if a proof can be found from there.
This is inherently limited by the model's proof search capability as well as the search budget, as falling short in either category could yield false negatives, but it is useful nonetheless for reducing noise.
Conversely, suppose we annotate $t$ as a positive tactic because we found a proof $\tau$ starting from $s$.
This could still be a false positive if $\tau$ immediately undoes $t$, returning to state $s'$ (i.e. $t$ ultimately did not contribute to the proof).
There are some countermeasures that depend on tracking state depths, but we leave a more complete solution to future work.

\begin{table*}[ht]
\centering
\begin{tabular}{l c c}
\hline
Reward Model & Llemma-7B Prompt & DeepSeek-Prover Prompt \\ \hline
SFT          & 86.6\%          & 87.9\%            \\ 
DPO          & 87.3\%          & 86.4\%            \\ \hline
\end{tabular}
\caption{\textbf{Reward Model Evaluation Comparison.} The table displays the accuracy of reward models in evaluation based on the fine-tuning method (SFT vs DPO) and the formatting for prompting (Llemma vs DeepSeek-Prover). The base model for these experiment is DeepSeek-Prover v1}
\label{tab:reward_model_comparison}
\end{table*}

Early results from SFT and DPO training yield mixed results.
We found the relative performance of the methods to be sensitive to adjustments in the prompt format.
We attribute the inconsistency to the low amount of exploration we used in annotating the training set and are currently preparing a scaled up version.

\subsection{Further Baselines}
For the early results, we primarily compare GFlowNet fine-tuning's efficacy to SFT.
A perhaps more comparable baseline still remains: Proximal Policy Optimization (PPO) \citep{schulman_proximal_2017}. As one of the most widely used forms of reward-maximizing reinforcement learning, we are working to implement a PPO training for the NTP setting. 

Proximal Policy Optimization seeks to optimize an objective that balances the trade-off between exploration and exploitation of the current policy \(\pi_{\text{old}}\) to maximize rewards and discover new policies \(\pi_{\theta}\). This balance is maintained using clipped probability ratios, which prevent updates from straying too far from the current policy. The standard PPO objective is expressed as:

\[
L(\theta) = \mathbb{E}_{s_t, a_t \sim \pi_{\text{old}}} \left[ \min \left( \frac{\pi_{\theta}(a_t|s_t)}{\pi_{\text{old}}(a_t|s_t)} \hat{A}(s_t, a_t), \ \text{clip} \left( \frac{\pi_{\theta}(a_t|s_t)}{\pi_{\text{old}}(a_t|s_t)}, 1 - \epsilon, 1 + \epsilon \right) \hat{A}(s_t, a_t) \right) \right],
\]

where \(\epsilon\) is a hyperparameter that sets the clipping range, and \(\mathbb{E}_{s_t, a_t}\) refers to the expectation over an on-policy batch of samples. The state-value function \(V(s)\) estimates the expected cumulative reward an agent can obtain starting from state \(s\), assuming it follows a given policy \(\pi\), which maps states to actions. The action-value function, or Q-function \(Q(s, a)\), estimates the expected cumulative reward starting from state \(s\), taking action \(a\), and then following the policy \(\pi\). \(\hat{A}(s_t, a_t)\) represents the advantage function, which is calculated as \(A(s, a) = Q(s, a) - V(s)\).

\end{document}